\documentclass[letterpaper, 10 pt, journal, twoside]{IEEEtran}  

\IEEEoverridecommandlockouts                              

\usepackage{graphicx}
\usepackage{amsmath}
\usepackage{amssymb}
\usepackage{booktabs}
\usepackage{adjustbox}
\usepackage[dvipsnames]{xcolor}
\usepackage{tabularx}
\usepackage{subcaption}
\usepackage{harpoon}
\usepackage{threeparttable} 
\usepackage{hyperref} 
\usepackage{pstool}

\usepackage{multirow}

\usepackage{cite}

\usepackage[capitalize]{cleveref}
\crefname{section}{Sec.}{Secs.}
\Crefname{section}{Section}{Sections}
\Crefname{table}{Table}{Tables}
\crefname{table}{Tab.}{Tabs.}

\definecolor{yb}{rgb}{0.855, 0.67, 0.102}
\definecolor{nb}{rgb}{0.3686274509803922, 0.5882352941176471, 0.8156862745098039}

\newcommand*{\V}{\mathbf}

\title{
	An Algorithm for the SE(3)-Transformation on Neural Implicit Maps for Remapping Functions
}

\author{Yijun Yuan and Andreas N\"uchter
\thanks{Manuscript received: February, 23, 2022; Revised May, 24, 2022; Accepted June, 15, 2022.}
\thanks{This paper was recommended for publication by Editor Javier Civera upon evaluation of the Associate Editor and Reviewers' comments.}
\thanks{The authors are with Informatics VII : Robotics and Telematics, University of W{\"u}rzburg
        {\tt\small \{yijun.yuan|andreas. nuechter\}@uni-wuerzburg.de }}%
\thanks{Digital Object Identifier (DOI): see top of this page.}
}

\begin{document}

\maketitle
\markboth{IEEE Robotics and Automation Letters. Preprint Version. Accepted June, 2022}
{Yuan \MakeLowercase{\textit{et al.}}: Transformation on Neural Implicit Maps} 
\begin{abstract}
Implicit representations are widely used for object reconstruction due to their efficiency and flexibility.
In 2021, a novel structure named neural implicit map has been invented for incremental reconstruction.
A neural implicit map alleviates the problem of inefficient memory cost of previous online 3D dense reconstruction while producing better quality.
However, the neural implicit map suffers the limitation that it does not support remapping as the frames of scans are encoded into a deep prior after generating the neural implicit map.
This means, that neither this generation process is invertible, nor a deep prior is transformable.
The non-remappable property makes it not possible to apply loop-closure techniques.
We present a neural implicit map based transformation algorithm to fill this gap.
As our neural implicit map is transformable, our model supports remapping for this special map of latent features.
Experiments show that our remapping module is capable to well-transform neural implicit maps to new poses.
Embedded into a SLAM framework, our mapping model is able to tackle the remapping of loop closures and demonstrates high-quality surface reconstruction. 
Our implementation is available at github\footnote{\url{https://github.com/Jarrome/IMT_Mapping}} for the research community.
\end{abstract}
\section{Introduction}

\IEEEPARstart{T}{he} reconstruction of 3D models has been explored for decades.
Its developments followed the trend of low-cost, high-quality sensors, and efficient computation hardware and were boosted in recent years with the progress in Deep Learning.

In the field of reconstruction, most attention has been drawn to global optimizations with bundle adjustment and loop closure~\cite{cao2018real,dai2017bundlefusion,whelan2015elasticfusion}.
Reconstructions using the Signed Distance Function (SDF) as a representation~\cite{curless1996volumetric} have been widely accepted as a fundamental basis since Kinect Fusion~\cite{newcombe2011kinectfusion} and VoxelHashing~\cite{niessner2013real}.
Recently, this basis is being challenged by the new trend of Deep Learning, as those conventional approaches have issues with the memory requirements and quality of uncomplete scans.

Relying on the high modeling ability of deep learning models, DeepSDF~\cite{park2019deepsdf} and Occupancy Networks~\cite{mescheder2019occupancy} propose implicit geometric representations that represent the shape in continuous space and thus are able to extract maps at an arbitrary resolution.
Similar to deep local descriptor~\cite{yuan2021self} that uses a parametric function to encode the geometry, the deep implicit model further supports prediction of the fields.
These deep implicit representations have been widely sought after for their flexible use in shape reconstruction~\cite{park2019deepsdf}, shape generation~\cite{chen2019learning} and more general tasks.
By operating on the voxel-level, \cite{chabra2020deep,jiang2020local} even provide semantics-agnostic high-quality reconstructions.

Relying on the success of deep implicit representation, in 2021, DI-Fusion~\cite{huang2021di} and NeuralBlox~\cite{lionar2021neuralblox} firstly proposed incremental neural implicit maps for 3D reconstructions.
DI-Fusion especially has introduced a novel reconstruction pipeline that effectively combined the advantage of the efficiency of neural implicit representation and the robustness of field base registration.
Different from the DI-Fusion that uses still the conventional SLAM pipeline, \cite{sucar2021imap, zhu2021nice} have proposed live-optimization with implicit representation for reconstruction which is also able to complete unseen surfaces.

However, there is still a severe limitation for an implicit representation compared with the common (point cloud, TSDF) methods:
implicit representations do not support transformation.
And it is this limitation that makes it hard to implement relocalization and remapping for neural implicit map reconstructions. Yuan et al. proposes indirect registration to evade the transformation of field during registration~\cite{yuan2022indirect}. However, the rotation and translation is inevitable for remapping function. 

\begin{figure}[t!]
	\centering
	\psfragfig[width=1\linewidth]{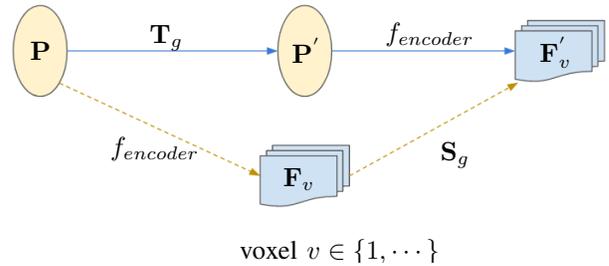}{
		\psfrag{A}{$\V T_g$}
		\psfrag{B}{$f_{encoder}$}
		\psfrag{C}{$f_{encoder}$}
		\psfrag{D}{$\V S_g$}
		\psfrag{E}{voxel $v \in \{1,\cdots\}$}
		\psfrag{P}{$ \V P$}
		\psfrag{Pb}{$\V  P^{'}$}
		\psfrag{Fb}{$\V  F_v^{'}$}
		\psfrag{F}{$\V  F_v$}
	}
	\caption{Two flow paths to SE(3)-transform and deep encode the point cloud. The {\color{nb}solid line} indicates the transform-encoding path to generate implicit map of $\V T_g$-transformed point cloud $\V P$. The {\color{yb}dash line} shows the encoding-transform path, transforming the map of features with transformation $\V S_g$, that is introduced in this paper.}
	\label{fig:twobranch}
	\vspace{-.4cm}
\end{figure}

In this paper, we propose a transformation algorithm for neural implicit maps to fill in this gap.
As shown in \cref{fig:twobranch}, encoding the transformed point cloud is equivalent to first encoding the points and transforming then on the neural implicit.
$\V S_g$ is the transformation on neural implicit corresponding to the Euclidean space transformation $\V T_g$.

The main challenge is the feature space transform with the corresponding alignment of the original point set.
Thus, in this paper, we exploit the topic of equivariant representation~\cite{thomas2018tensor,fuchs2020se,deng2021vector}, to implement the implicit map transformation. 
As the focus of SE(3)-equivariant research is not the transformation,
it is actually not adequate to solve full transformation in feature space.
In addition, the recent approaches work only on small examples with simple structures.
Thus, our proposed model works on a map of neural implicits, i.e., a set of neural implicit functions, instead of one holistic implicit function, to evade both limitations.

The transformed implicit map is able to produce a very close result to the implicit map of the transformed point cloud.
To demonstrate the advantage of our mapping model, we also embed it into a loop-closure equipped SLAM-algorithm~\cite{mur2017orb}.

The contributions of this paper are as follows:
\begin{itemize}
	\item We propose a transformation algorithm for neural implicit maps.
	\item We implement a 3D reconstruction with this remapping model.
\end{itemize}

In the following, we first describe briefly the related work for implicit function and equivariant features.
Then, we introduce our transformation algorithm for neural implicit maps.
After that, experiments demonstrate the performance and we conclude this work.

\section{Related Work}

\subsection{Deep Implicit Representations}

Algorithms for Implicit Representation can be divided into two categories:
First, the most widely used branch is the DeepSDF~\cite{park2019deepsdf}.
For SDF, the geometry prior is encoded with MLP and then fed into another model together with query points to extract the signed distance values with a discretized distance field.
A mesh is then extracted using the Marching Cubes algorithm~\cite{lorensen1987marching}.
For non-closed shapes which is more general for point cloud data, an unsigned distance field neural model is introduced without indicating inside-outside~\cite{chibane2020neural}.
The second category is Occupancy Networks~\cite{mescheder2019occupancy}.
For Occupancy Networks, different from the distances in the SDF model, the probability of occupancy at a certain position is estimated from the implicit function.
Then a Multiresolution IsoSurface Extraction (MISE) is implemented to obtain meshes.

To efficiently reconstruct intricate surfaces, DeepLS~\cite{chabra2020deep} introduces a local deep geometry prior and performs the reconstruction with a set of local learned continuous SDFs.
Similarly, \cite{jiang2020local} proposes a local implicit grid for reconstruction.
Note that, one advantage of the local implicit model is that it relieves the pressure on the encoding model.
As a local surface is much more simple compared to a whole complicated scene, such a local strategy is trained on a simple synthetic object dataset and generalized then to the real complex scene.

In 2021, DI-Fusion~\cite{huang2021di} moves one step further and leads the research to a real reconstruction of a scene.
It is the first implicit function research that realizes the incremental reconstruction of a scene.
More importantly, they alleviate the memory inefficiency of SDF representation by updating a map of latent features while extracting distance values for registration and visualization, yielding a new direction of 3D reconstruction with Deep Learning.
Similarly, NeuralBlox~\cite{lionar2021neuralblox} also proposes to fuse the grid of latent features. Given an external state estimation with noise, its latent code fusion still shows a robust performance.
iMAP~\cite{sucar2021imap} proposes a non-conventional SLAM pipeline with implicit neural representation for incremental reconstruction.
The main component of it is a differentiable rendering model.
With an online optimization, the reconstruction is optimized by repeatedly minimizing the rendering distance to observed images.

\begin{figure}[!]
	\centering
	\includegraphics[width=.8\linewidth]{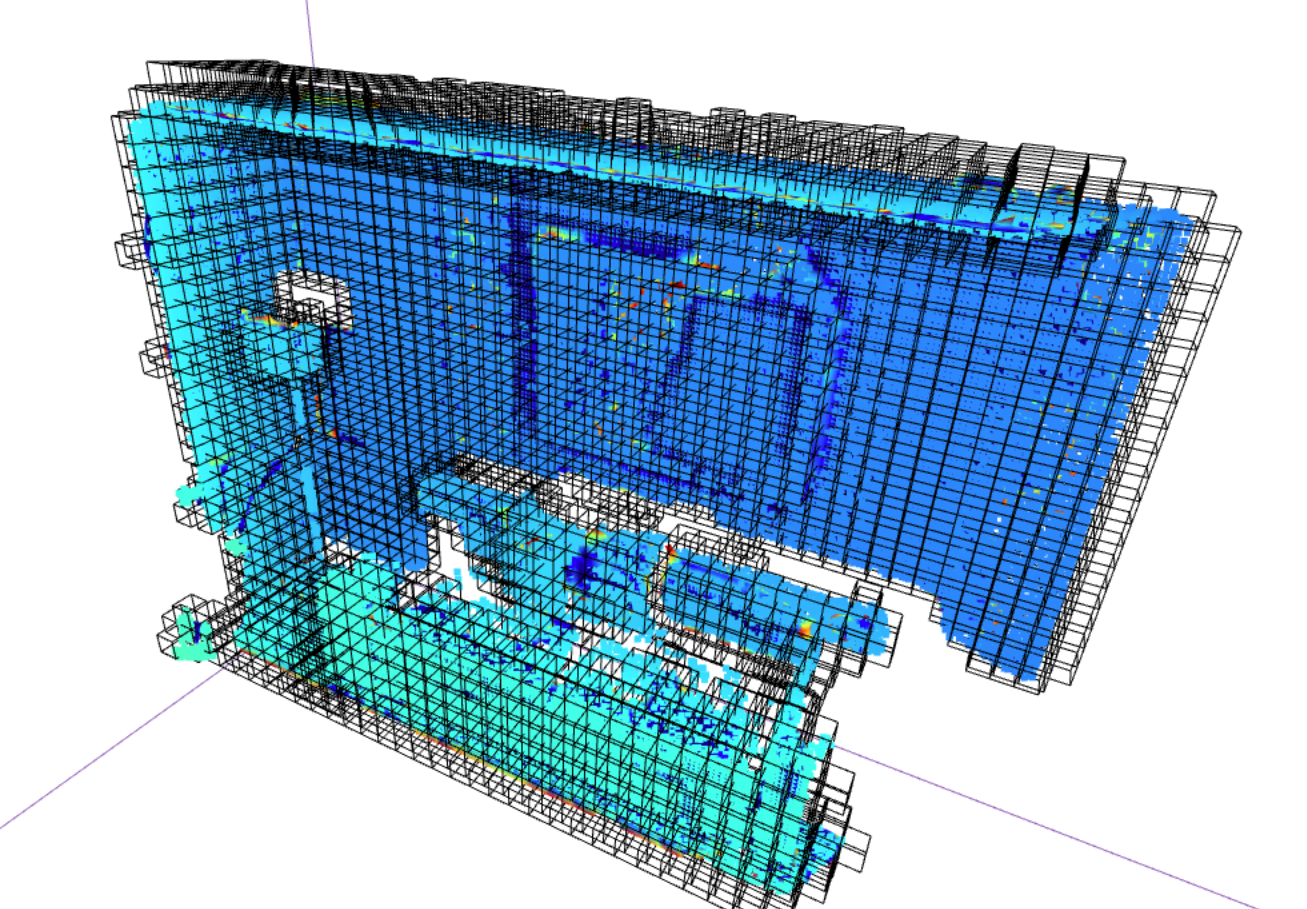}
	\caption{PLIVox representation from DI-Fusion~\cite{huang2021di}.}
	\label{fig:PLIVox}
	\vspace{-0.5cm}
\end{figure}

In this work, we build on top of DI-Fusion by using their PLIVox
representation as in \cref{fig:PLIVox}.

\subsection{Equivariant Feature}
Equivariance is a novel concept for 3D point clouds.
The target is a universal representation of objects with different poses to avoid exhaustive data augmentation. 

We follow the definition in SE(3)-Transformers~\cite{fuchs2020se}:
Given a set of transformations $\V  T_g: \mathcal V\rightarrow \mathcal V$ for $g \in G$, where G is an abstract group, a function $\phi:\mathcal V \rightarrow \mathcal Y$ is equivariant if for $g$, there exists a transformation $\V S_g: \mathcal Y \rightarrow \mathcal Y$ such that 
\begin{align}
 \V S_g[\phi(v)] = \phi( \V T_g[v]) \text{\ \ for all\ } g\in G,v\in \mathcal V.
\end{align}


\begin{figure}[b!]
	\vspace{-0.5cm}
	\centering
	\psfragfig[width=1\linewidth]{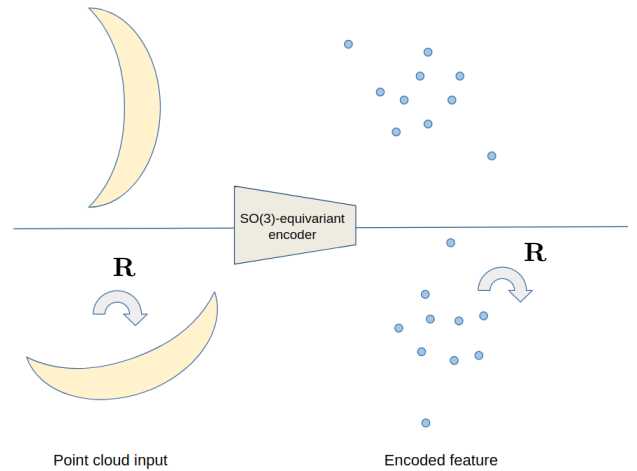}{
		\psfrag{R}{$\V R$}
	}
	\caption{SO(3)-equivariant representation for point cloud.}
	\label{fig:SO3}
\end{figure}
\begin{figure*}[]
	\centering
	\includegraphics[width=.9\linewidth]{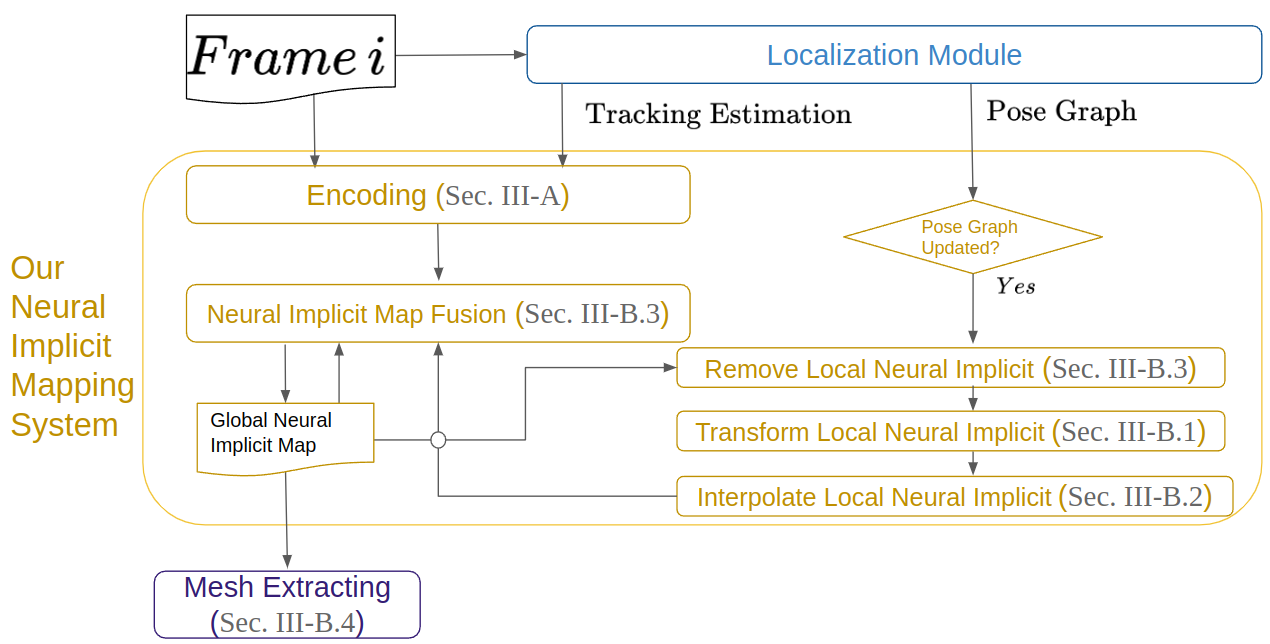}
	\caption{Pipeline for SLAM embedding our mapping module. {\color{nb}SLAM module} provides the point cloud $P_i^{'}$ and the pose table $\{T_1,\cdots,T_i\}$ for the {\color{yb} mapping module} with keyframe $i$.}
	\label{fig:pipeline}
	\vspace{-0.5cm}
\end{figure*}


From their definition, an SO(3)-equivariant function $\phi$ follows $\V S_{\V R}\phi(v) = \phi(\V R[v])$ with function $\V S$ an operation to produce same result as aligning the point cloud.
While for the translation-equivariant, for convenience, the $\V S$ operation is usually defined as identity mapping. 

As the translation is reduced with the relative position, most of works mainly focus on SO(3)-equivariant representations~\cite{thomas2018tensor, kondor2018clebsch, esteves2018learning, weiler20183d} with steerable kernel bases. SE(3)-Transformers~\cite{fuchs2020se} leverages the advance of self-attention on large point sets and graphs with various point numbers.  

Realizing equivariance by learning, those works are restricted to using convolutions and relative positions of neighboring points.
Vector Neurons (VNN) firstly introduce a whole group of network layers that produce SO(3)-equivariant features~\cite{deng2021vector}.
It is flexible to reconstruct PointNet~\cite{qi2017pointnet} or DGCNN~\cite{phan2018dgcnn} with VNN layers.
This provides us a good basis as it is able to function as point-encoder. In this work, we mainly use three rotation-equivariant operations: VNLinear, VNLeaKyReLU, and mean-pooling. For example, with the input $\V V\in\mathbb{R}^{C\times 3}$, VNLinear parameter $ \V W_l\in\mathbb{R}^{C^{'}\times C}$, VNLinear produces the output $\mathbf W_l\mathbf V$ where $(\V W_l\V V)\V{R}=\V W_l(\V V\V R)$ is rotation-equivariant. 
LeaKyReLU produces each output vector-neuron $\V v^{'}_c\in \V V^{'}=f_{\text{LeaKyReLU}}(\V V)$ with separate parameters $\V W_c\in \mathbb{R}^{1\times C}$ and $\V U_c\in \mathbb{R}^{1\times C}$ where $c\in \{1,\cdots, C\}$. 
For each vector-neuron $\V v^{'}_c\in \mathbb{R}^{1\times 3} $, it maps the input feature $\V V$ 
to the feature $\V q_c = \V W_c \V V\in \mathbb{R}^{1\times 3}$ and direction $\V k_c = \V U_c \V V \in \mathbb{R}^{1\times 3}$. 
Thus it produces
\begin{equation}
\V v^{'}
= \begin{cases}
\V q_c &
\text{if\ } \langle \V q_c,\V k_c \rangle \geqslant 0 \\	
\V q_c - \left\langle \V q_c,\frac{\V k_c}{\|\V k_c\|} \right\rangle \frac{\V k_c}{\|\V k_c\|} &
\text{otherwise,}
\end{cases}
\label{eq:VN-ReLU}
\end{equation}
with the output $\V V^{'}=[\V v^{'}_c]_{c=1}^C$~\cite{deng2021vector}. Mean-pooling is an average on the same dimension of all points, therefore it is naturally rotation-equivariant.

As the original goal of the equivariance concept is to provide universal features, i.e., $\V S=\V I$ is adequate for translation.
However, with a different focus of transforming the feature space, this setting is not applicable.
In this paper, only the feature rotation resorts to the SO(3)-equivariant architecture VNN~\cite{deng2021vector} and functions as \cref{fig:SO3}.
The translation is solved using other techniques.

\section{Methodology}

We follow DI-Fusion~\cite{huang2021di} to use the evenly-spaced voxels (PLIVoxs) to represent the map.
$\V  V = \{\V  v_m=(\V c_m, \V F_m, w_m) \}$ with $\V c_m \in \mathbb R^3$, $\V F_m\in \mathbb R^{L}$, $w\in \mathbb N$ the voxel centriod, latent representation of observed geometry and observation count respectively.
\subsection{SO(3)-equivariant Features}
\label{sec::feature}
%

\begin{figure}[t!]
	\centering
	\psfragfig[width=.9\linewidth]{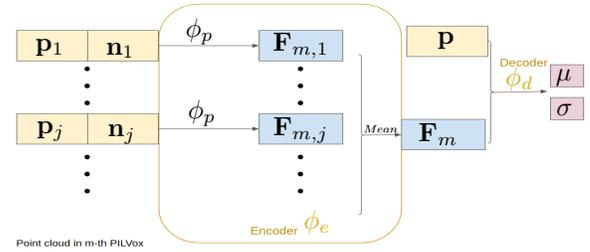}{
		\psfrag{p1}{$\V p_1$}
		\psfrag{n1}{$\V n_1$}
		\psfrag{pj}{$\V p_j$}
		\psfrag{nj}{$\V n_j$}
		\psfrag{phip}{$\phi_p$}
		\psfrag{Fm1}{$\V F_{m,1}$}
		\psfrag{Fmj}{$\V F_{m,j}$}
		\psfrag{phie}{$\color{yb}{\phi_e}$}
		\psfrag{phed}{$\color{yb}{\phi_d}$}
		\psfrag{Fm}{$\V F_{m}$}
		\psfrag{p}{$\V p$}
		\psfrag{m}{$\mu$}
		\psfrag{s}{$\sigma$}
	}
	\caption{The structure of encoder-decoder neural network.  $\V {p_j}$, $\V {n_j}$ are point xyz and norm for certain point $\V p_j$ in one $m$-th PLIVox. $\V p$ are point for inference and $\mu$, $\sigma$ are estimated distance value and its standard derivation.}
	\label{fig:point_encoder}
\end{figure}

Our encoder-decoder neural network $\Phi$ follows the design of encoder-decoder in DI-Fusion~\cite{huang2021di}.
$\phi_e$ encodes points in a PLIVox, and $\phi_d$ predicts distance mean and standard deviation for query points.
But different from DI-Fusion that is using a simple PointNet structure, to realize the transformation on the neural implicit map, we propose to use VNN layers~\cite{deng2021vector} to build an SO(3)-equivariant encoder.

For each local voxel, it uses the points $\V P_m$ and the norm $\V S_m$ as an input.
Two branches of VNN-MLPs are respectively applied on $\V P_m$ and $\V S_m$ and produce features $\V F_{\V P_m}\in \mathbb R^{l\times 3}$ and $\V F_{\V S_m} \in \mathbb R^{l\times 3}$.
Then by concatenating $\V F_{\V P_m}$ and $\V F_{\V S_m}$ along the $l$ axis, it achieves $\V F_m\in \mathbb R^{2l\times 3}$.
A point encoder $\phi_p$ is given in \cref{fig:point_encoder}.
The local point set encoder $\phi_e$ produces the mean-pooling of the $\phi_p$ output in $\V P_m$. 

In this encoder-decoder we changed the encoder with VNN to realize the SO(3)-equivariant functionality.
For more details about the decoder network and Conditional Neural Processes-style training, please refer to DI-Fusion~\cite{huang2021di}.

\subsection{Neural Implicit Mapping Module}

Neural Implicit Mapping Module consists of Encoding~(\cref{sec::feature}), Fusion~(\cref{sec:map_removal_fusion}), Removal~(\cref{sec:map_removal_fusion}), and Transforming~(\cref{sec::map::trans}) functions. 

The input frame to this module is firstly encoded into a local neural implicit map and fused to the global map.
When a loop is detected, remapping a certain frame requires removing, transforming, and fusing the corresponding local neural implicit map.

A diagram of our mapping module is given in \cref{fig:pipeline}.
The neural implicit mapping module serves as a mapping module for SLAM.

\subsubsection{Transformation to Global Coordinate}
\label{sec::map::trans}

Our transformation algorithm of Neural Implicit Map consists of two steps: the grid transformation and the feature rotation.

\begin{figure}[b!]
	\centering
	\psfragfig[width=1.\linewidth]{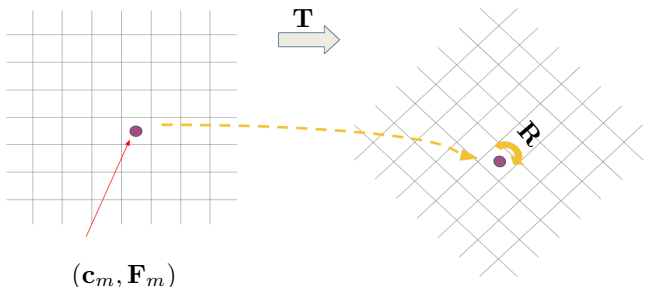}{
		\psfrag{R}{$\V R$}
		\psfrag{F}{$(\V c_m, \V F_m)$}
		\psfrag{T}{$\V T$}
	}
	\caption{Demonstration of the transformation on the neural implicit map. The voxel grid is transformed to a new position. $\V F_m$ rotates since $\V F_m$ is still positioned at the center of a voxel in the grid after the transformation (center is transformed).}
	\label{fig:transform}
\end{figure}

As demonstrated in \cref{fig:transform},
given the transformation $\V T$ of the local map $\V V_l$ to global coordinates, the update is actually on the center $\V c$ and its corresponding implicit feature $\V F$ for each PLIVox.
For PLIVox voxel $\V v_m\in \V V_l$, center (grid coordinate) $\V c_m$ directly transforms
\begin{align}
\V c_m \leftarrow \V T\cdot \V c_m.
\end{align} 
Afterwards, for the feature $\V F_m\in \mathbb{R}^{2l\times3}$, as the feature is always positioned at the voxel center, the rotation is left to solve. Thus
\begin{align}
\V F_m \leftarrow ( \V R \cdot \V F_m^{T} )^T.
\label{eq:feature_transform}
\end{align} 
However, only transforming the local neural implicit frame is not sufficient to update the global map. Because the transformed voxel grid for the local frame may not be consistent with the grid of the global map.

\subsubsection{Interpolation to Global Grid}
\label{sec::map::interp}

\begin{figure}[b!]
	\vspace{-0.6cm}
	\centering
	\psfragfig[width=.6\linewidth]{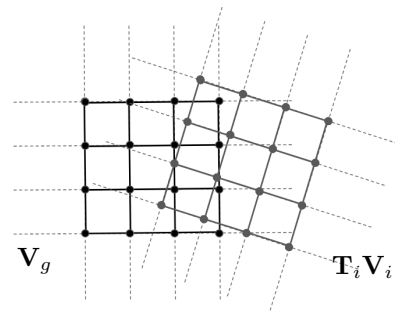}{
		\psfrag{T}{$\V V_g$}
		\psfrag{V}{$\V T_i \V V_i$}
	}
	\caption{The transformed local grid is not well-fitted with the global grid. Here we show the center of the grid for better demonstration.}
	\label{fig:transformed}
\end{figure}

As shown in \cref{fig:transformed}, there is a small gap between global and local grid coordinates.
Therefore, we need to additionally interpolate the local features on the global grid.
%

Since the distance between the local and the target voxel is small, and points involved for encoding are actually in a $2$-times voxel length region around its voxel in implementation, we propose to align voxels by linearizing the function $\phi_e(\V P+\V t)$.
Each $\V v_m$ in the local grid will contribute to the close neighbor $\V v_{n}$ in the global grid:
\begin{align}
	\V F_{\V v_{n}} = \phi_e(\V P_m + \V t_{m,n} ), \ \ \ \V  t_{m,n} = \V c_{n}-\V c_m
	\label{eq:f_nc}
\end{align}

Then by linearizing the right side of \cref{eq:f_nc}, we yield
\begin{align}
	\V F_{\V v_{n}} = \V F_{\V v_m} + \frac{\partial}{\partial \V t} [\phi_e(\V P_m  )] \V t_{m,n}.
	\label{eq:f_lin}
\end{align}
Here we denote the Jacobian as $\V{J}_m = \frac{\partial}{\partial \V t} [\phi_e(\V P_m  )]$, for which $\V J_m\in \mathbb{R}^{2l\times3\times3}$.

Following the feature metric of PointNetLK~\cite{aoki2019pointnetlk}, we approximate each column of the Jacobian using 
\begin{align}
	\V J_{m,p}\approx \frac{ \phi_e( \V P_m+\V t_p)-\phi_e(\V P_m)}{\Delta t} \in \mathbb{R}^{2l\times 3}
	\label{eq:Jacobian_1}
\end{align}
where $\V t_p \in \{[\Delta t,0,0],[0,\Delta t,0], [0,0,\Delta t]\}$. 

Then the Jacobian of the feature over the translation is
\begin{align}
	\V J_m = [\V J_{m,1}\    \V{J}_{m,2} \  \V {J}_{m,3}].
	\label{eq:Jacobian_2}
\end{align} 
To note that the Jacobian is computed together with the implicit feature which is \emph{before} the transformation (with $\V T$).
Thus each column of Jacobian need a pre-transformation $\V J_{m,p}\leftarrow \V J_{m,p}\V R^T$ together with its feature transformation in \cref{eq:feature_transform}.
In addition, the translation bias $\V t_{m,n}$ in \cref{eq:f_lin} cannot be directly multiplied with $\V J_m$.
An inverse rotation is required, that is $\V J_m \V R^T \V t_{m,n}$.
We then keep the formulation \cref{eq:f_lin} still valid by rewriting the Jacobian as
$\V J_m\leftarrow \V J_m \V R^T$.

In our implementation, for each target grid $\V v_n$ in $\V V_g$ that has neighbors with center distance$<$voxelSize, we find its $K$ nearest neighbors $\V v_m$ where $m\in\{c_1,\cdots c_K \}$ in the local grid. Here $c_i$ denotes the PLIVox index of neighbors.
Then we interpolate
\begin{align}
	\V F_{n} = \sum_{m\in\{c_1,\cdots c_K \}} s_{n,m}( \V F_m + \V J_m   \V t_{n,m} )
\end{align}
where $s_{n,m} = \frac{\exp(-||\V t_{n,m}||^2)}{\sum\exp( -|| \V t_{n,\cdot}||^2)}$.
Moreover, the voxel point number $w$ is required for the following global neural implicit map updating (\ref{sec:map_removal_fusion}), thus should also be recorded
\begin{align}
	w_{n} = \sum_m s_{n,m} \cdot w_{m}.
\end{align}

\subsubsection{Map Removal \& Fusion}
\label{sec:map_removal_fusion}

DI-Fusion provided an updating of the neural implicit map in a voxel-to-voxel manner.
As we transform and fit the local grid to the global grid in previous \cref{sec::map::trans}~\cref{sec::map::interp}, we are now ready for the neural implicit map updating.

Since the global map has been fused previously with the local map $\V V_{k}$, i.e., after a pose update, the global map $\V V_g$ requires a local removal, and afterwards a local fusion with the updated neural implicit map. 
We similarly formulate the removal formula.
The removal and fusion are done as following:
\begin{align}
	\V F_m\leftarrow \frac{\V F_mw_m\mp \V F_m^k}{w_m \mp w_m^k},\quad w_m\leftarrow w_m\mp w_m^k
\end{align}

\subsubsection{Mesh Extracting}

We follow DI-Fusion~\cite{huang2021di} to build the reconstruction model from the neural implicit map. 
First, signed distance fields are generated for each PLIVox by using the decoder $\phi_d$ from \cref{sec::feature}.
Then with the one complete signed distance field, the Marching Cube algorithm is used to extract the mesh model.

The whole pipeline is plotted in \cref{fig:pipeline}.
When frame $i$ is processed by the SLAM system, an external localization module (see experiment section) is required to track the camera and maintain the pose graph.
In each frame, our neural implicit mapping module encodes the frame and fuses it into the global implicit representation.
When there is a loop closure, it updates the sequence of poses and our mapping system checks the pose of each frame.
If the pose of a certain frame is updated, the mapping module will \emph{(1) remove the old local neural implicit map} of that frame, \emph{(2) transform to a new pose and interpolate} to produce a new local neural implicit map from the original copy of the local map, and \emph{(3) fuse this new local map into global}.


\section{Experiments}


\subsection{Setting}
\subsubsection{Datasets}

Three datasets have been used in our experiments.
The object dataset ShapeNet~\cite{chang2015shapenet} is used for training purpose.
The RGB-D dataset ICL-NUIM~\cite{handa2014benchmark} and 
Replica~\cite{straub2019replica} are utilized for quantitative evaluation. 


\paragraph{ShapeNet~\cite{chang2015shapenet}}
ShapeNet is a rich-annotated large variety 3D shape dataset. 
We follow \cite{huang2021di} to select 6 categories (bookshelf, display, sofa, chair, lamp, and table) and 100 samples to train the encoder and decoder model. 
For more details about the pre-procession of this data, please refers to \cite{huang2021di}.

\paragraph{ICL-NUIM~\cite{handa2014benchmark}}
ICL-NUIM is a widely used RGB-D dataset for SLAM and Reconstruction. 
It contains living rooms and office room scenes. 
From which the living room scene contains a ground truth surface model. So this living room scene is widely involved in research for surface comparison. 
We use lr-kt[0-3] with synthetic noise for standard surface comparison.

\paragraph{Replica~\cite{straub2019replica}}
The Replica data set is a highly photo-realistic 3D indoor scene reconstruction dataset. 
We use iMAP's~\cite{sucar2021imap} 8 sequences (5 offices and 3 apartments) from Replica. 
The 8 sequences contain rendered 2000 RGB-D frames each. 
Different from ICL-NUIM sequences that do not repeatedly record certain views, because of the live-optimization of iMAP, the replica sequences cover each direction and surface multiple times.
We extensively implement our model on this dataset to demonstrate the reconstruction quality.

%

\subsubsection{Implementation details}

All of the experiment is implemented on a NUC-computer (CPU-i7-10710U 1.10GHz, 32GB memory, GTX2080Ti-12GB). 
We follow the DI-Fusion to set PLIVox parameters: voxel-size$=0.1m$.
For mesh extraction, we also set the same $\sigma$ threshold $\sigma_D=0.06$ to fairly compare with DI-Fusion.
For encoding, we set the length of a feature to $9\times 3\times2$ for the VNN SO(3)-equivariant feature.
Three VNN-linear operations are required on each point and normal to get the two $9 \times 3$ features for each point in PLIVox. More specifically, the point encoder sequentially VNLinear(1,32)$\rightarrow$VNLeakyReLU$\rightarrow$VNLinear(32, 32)$\rightarrow$VNLeakyReLU$\rightarrow$VNLinear(32,9) results in a $9\times 3$ size feature at point and normal branch respectively.
Then, by mean-pooling and concatenating, one $18 \times 3$ size point set feature is obtained for that PLIVox.
For decoding and optimization loss, we follow the same as DI-Fusion to predict both mean and variance and train the whole encoder-decoder network in a similar strategy as the Conditional Neural Processes~\cite{garnelo2018conditional}.
For interpolation, we set the candidate number $K=8$.
For mesh extracting, resolution$=4$ for the space grid in each PLIVox used.

To test on the Replica dataset, we set $\sigma_D=0.15$ and resolution$=3$.

\subsubsection{Training}

Even with the different encoder structures, our model is actually trained with the same setting as DI-Fusion~\cite{huang2021di} on ShapeNet data.
\begin{figure}[b!]
	\vspace{-0.5cm}
	\centering
	\includegraphics[width=.8\linewidth,height=.4\linewidth]{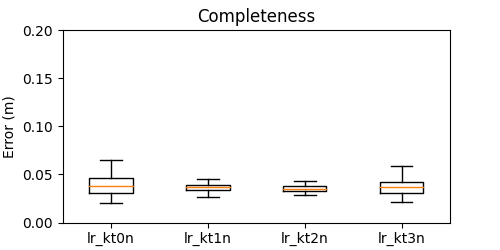}
	\includegraphics[width=.8\linewidth,height=.4\linewidth]{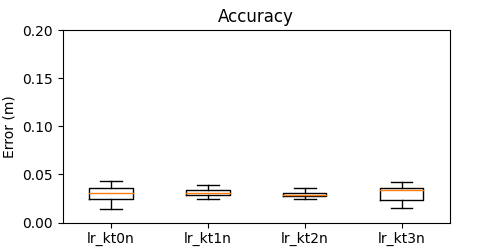}	
	\caption{Accuracy and completeness.
       Accuracy metrics show how close {\color{YellowOrange}encode-transform} extracted points to points in {\color{blue}transform-encode branch} are.
       While completeness metrics show how close {\color{blue}transform-encode} extracted points to points in {\color{YellowOrange}encode-transform branch} are.}
	\label{fig:exp:transform}
\end{figure}

\subsubsection{Testing}

During testing, the pre-trained encoder-decoder is transferred to the new scene of indoor scale reconstruction.
We have two tests, one is about the functionality of the transformation, and the other is about the incremental reconstruction.
For the first test in \cref{sec::exp::trans}, we use the mapping module for a single frame. 

For the second test in \cref{sec::exp::recons}, ORB-SLAM2 RGB-D is utilized to provide the localization.
Then we use two benchmarks to evaluate the performance: ICL-NUIM and Replica~\cite{straub2019replica}.
ICL-NUIM is the most widely used standard test which has a standard surface model and metrics tools for comparison. 

On the standard ICL-NUIM benchmark, we compare with DVO-SLAM~\cite{kerl2013dense}, Surfel Tracking~\cite{keller2013real}, ElasticFusion~\cite{whelan2015elasticfusion}, BundleFusion~\cite{dai2017bundlefusion}, and DI-Fusion~\cite{huang2021di}.

On Replica dataset, we follow the paper iMap~\cite{sucar2021imap} to select the data and compare it with iMAP as a baseline. 
Values are taken from the iMAP paper as its source is not released.

\subsection{Evaluate the Functionality of Transformation on Neural Implicit Maps}
\label{sec::exp::trans}

As introduced in \cref{fig:twobranch}, there are actually two paths to encode point clouds into the neural implicit map with transformation. 
To evaluate the functionality of our transformation algorithm, we generate neural implicit maps in both branches and then measure the performance with respect to accuracy and completeness between the reconstruction from the result neural implicit maps.
Accuracy shows the average distance between the sampled reconstruction points in the encode-transform path and the nearest points in the transform-encode path.
Completeness shows the average distance between the sampled points from transform-encode reconstruction and the nearest points in the encode-transform path.
We select lr-kt[0-3] as the test sequences, and GT-trajectory to provide the transformation accordingly. 
After generating the neural implicit maps, the decoder is used to generate the Signed Distance Field and the Marching Cubes algorithm is used to generate the surface. 
Each frame is recorded separately to compute the surface error. 
We also draw the error for each frame, the distribution of error is shown in \cref{fig:exp:transform}.
It is clear that our method retains a similar reconstruction with the transformation on the neural implicit maps.

For the completeness especially, the very low error shows that our transformation-on-implicit well-reconstructs the surface region compared to the transformation-then-implicit build.
%
However, there still exists an $\sim 3cm$ error.
The success of the incremental reconstruction in the following experiment shows that this is more a small surface generation effect in the whole reconstruction. 

\begin{table}[]
	{
		\small
		\centering
		\setlength{\tabcolsep}{6.0pt}
		\caption{Comparison of surface error on ICL-NUIM~\cite{handa2014benchmark} benchmark (measured in centimeters).}
		\label{tbl:icl-surface}
		\begin{tabularx}{\linewidth}{X|cccc} 
			\toprule
			& lr kt0        & lr kt1        & lr kt2        & lr kt3          \\ 
			\midrule
			DVO-SLAM~\cite{kerl2013dense}      & 3.2          & 6.1          & 11.9          & {5.3}           \\
			RGB-D SLAM~\cite{endres2012evaluation} & 4.4 & 3.2 & 3.1 & 16.7\\
			MRSMap~\cite{stuckler2014multi}& 6.1 & 14 & 9.8 & 24.8\\
			Kintinuous~\cite{whelan2015real} &1.1&0.8& 0.9& 15.0\\
			ElasticFusion~\cite{whelan2015elasticfusion} &0.7 & 0.7& \textbf{0.8}&2.8\\
			DI-Fusion~\cite{huang2021di}     & \textbf{0.6} & 1.5 & 1.1 & 4.5   \\
			
			\midrule
			Ours &1.2 & 1.58 & 1.0 & \textbf{1.2} \\	
			\bottomrule
		\end{tabularx}
	}
	\vspace{-0.4cm}
\end{table}
\subsection{Evaluate the Incremental Reconstruction Performance}
\label{sec::exp::recons}

\subsubsection{ICL-NUIM test}
\label{sec::exp::recons::ICL}
In this part, we evaluate our model on the ICL-NUIM benchmark with synthetic noise added. 
We use a surface error to metric for the difference between reconstruction and the ground-truth model.
The quantitative evaluation is demonstrated in \cref{tbl:icl-surface}.
We observe that the neural implicit map based algorithm achieves high accuracy compared to others.
However, in lr-kt3 which contains loops, DI-Fusion does not exceed ElasticFusion. But our model gets the best score with $1.2$.
To note that, our model is able to detect and remap the start-end loop on lr-kt3, which is reflected on the scores, $1.2cm$, exceeding the rest.
This demonstrates that our model is able to \emph{address the problem of DI-Fusion which is not compatible to a loop closure module}.
We find that our model scores similar on lr-kt1, 2 with DI-Fusion.
It also approves that our VNN-encoder well-represent the feature while holding the SO(3)-equivariant functionality.

\begin{table*}[!]
	\centering
	\caption{Reconstruction Test on Replica Dataset~\cite{straub2019replica}.}
	\label{tab:replica}
	\begin{adjustbox}{max width = \linewidth}
		\begin{threeparttable}
			\setlength{\tabcolsep}{0.36em}
			\begin{tabular}{clccccccccc}
				\toprule
				& & \tt{room-0} & \tt{room-1} & \tt{room-2}  & \tt{office-0} &  \tt{office-1} & \tt{office-2} & \tt{office-3} & \tt{office-4} & Avg. \\
				\midrule

				\multirow{3}{*}{iMAP$^*$~\cite{sucar2021imap} }   
				& {\bf Acc.} [cm] $\downarrow$
				& 3.58 & 3.69 & 4.68 & 5.87 & 3.71 & 4.81 & 4.27 & 4.83 & 4.43 \\
				& {\bf Comp.} [cm] $\downarrow$
				& 5.06 & 4.87 & 5.51 & \textbf{6.11} &\textbf{ 5.26}  & \textbf{5.65} & 5.45 & 6.59 &  \textbf{5.56}\\
				& {\bf Comp. Ratio} [$<$ 5cm \%] $\uparrow$
				& 83.91 & {83.45} & 75.53 & 77.71& \textbf{79.64} & 77.22& 77.34 & \textbf{77.63} & 79.06\\
  		
				\midrule		
				\multirow{3}{*}{\textbf{Ours}} 
				& {\bf Acc. } [cm] $\downarrow$ 
				&\textbf{ 2.05} & \textbf{1.74}&\textbf{1.97} & \textbf{2.03 }& \textbf{1.63} & \textbf{2.10 }& \textbf{2.75} & \textbf{3.07} & \textbf{2.17} \\
				& {\bf Comp. } [cm] $\downarrow$ 
				&\textbf{3.75}  &\textbf{3.41} &\textbf{4.60} & 9.68 & 8.73 & \textbf{5.67} & \textbf{4.77} & \textbf{5.14}  & 5.72 \\
				& {\bf Comp. Ratio } [$<$ 5cm \%] $\uparrow$ 
				& \textbf{86.59} & \textbf{87.60 }& \textbf{83.57}& \textbf{79.28} & 78.14 & \textbf{77.76} &\textbf{78.19}  & 74.16 & \textbf{80.66}  \\
				\bottomrule
			\end{tabular}%
			\begin{tablenotes}
				\item[$^*$] Values taken from \cite{sucar2021imap}.
			\end{tablenotes}
		\end{threeparttable}
	\end{adjustbox}
	\vspace*{-2mm}
\end{table*}

\begin{figure*}[t!]
	\centering
	\begin{subfigure}{.2\textwidth}
		\centering
		\includegraphics[width=1.\linewidth]{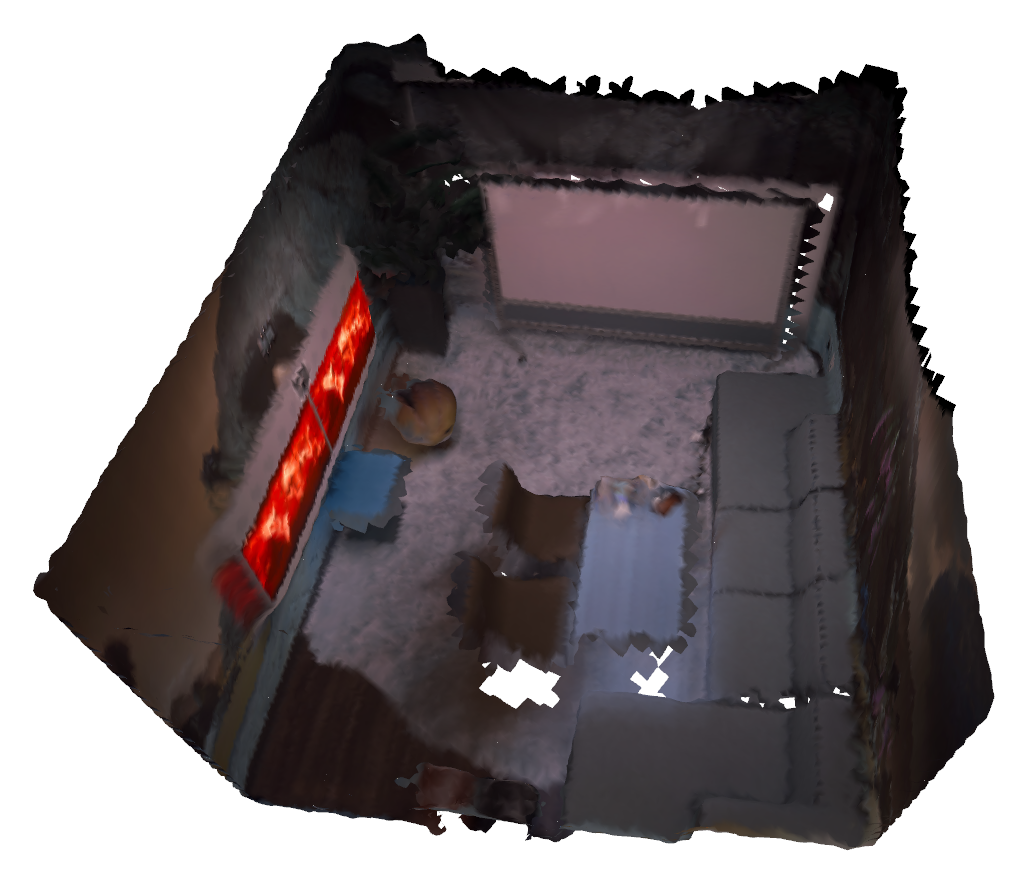}
		\caption{office0}
	\end{subfigure}%
	\begin{subfigure}{.2\textwidth}
		\centering
		\includegraphics[width=1.\linewidth]{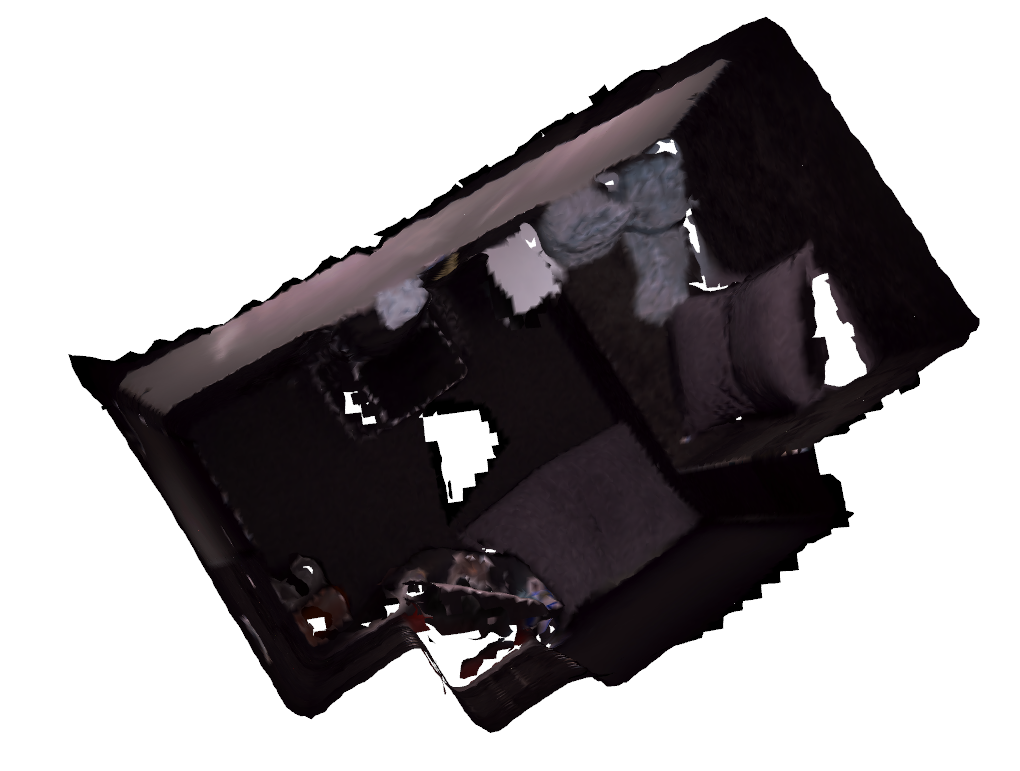}
		\caption{office1}
	\end{subfigure}%
	\begin{subfigure}{.2\textwidth}
		\centering
		\includegraphics[width=1.\linewidth]{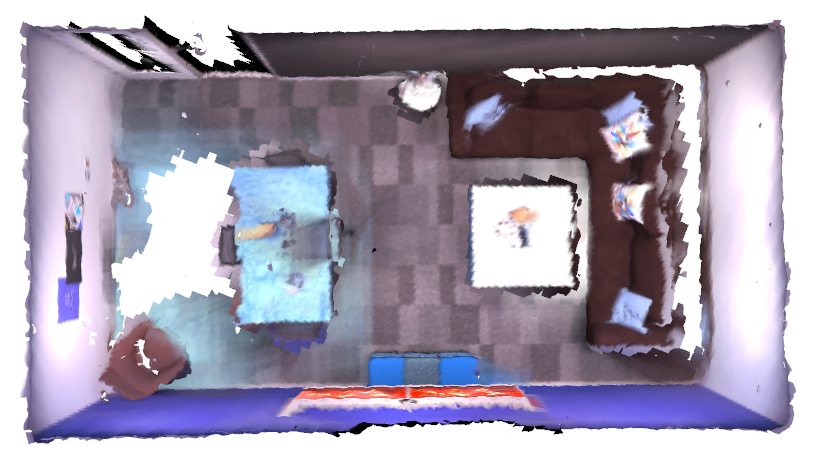}
		\caption{office2}
	\end{subfigure}%
	\begin{subfigure}{.2\textwidth}
		\centering
		\includegraphics[width=1.\linewidth]{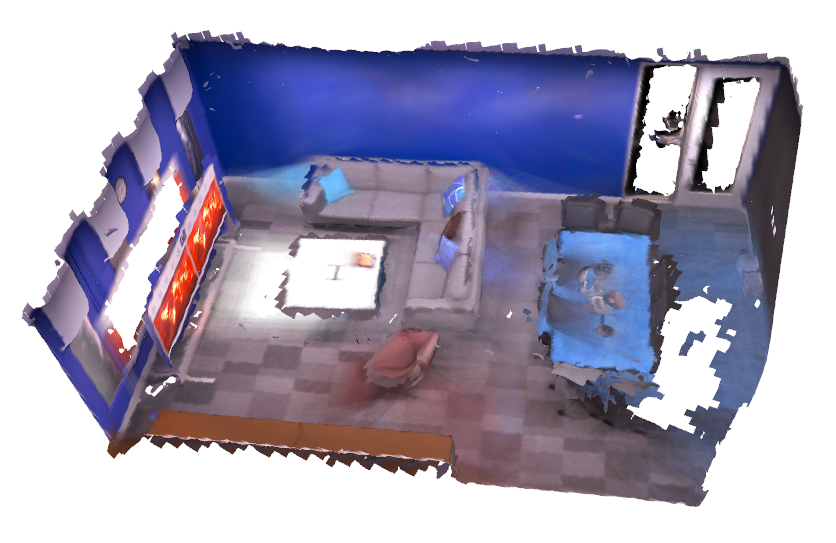}
		\caption{office3}
	\end{subfigure}%
\\
	\begin{subfigure}{.2\textwidth}
		\centering
		\includegraphics[width=1.\linewidth]{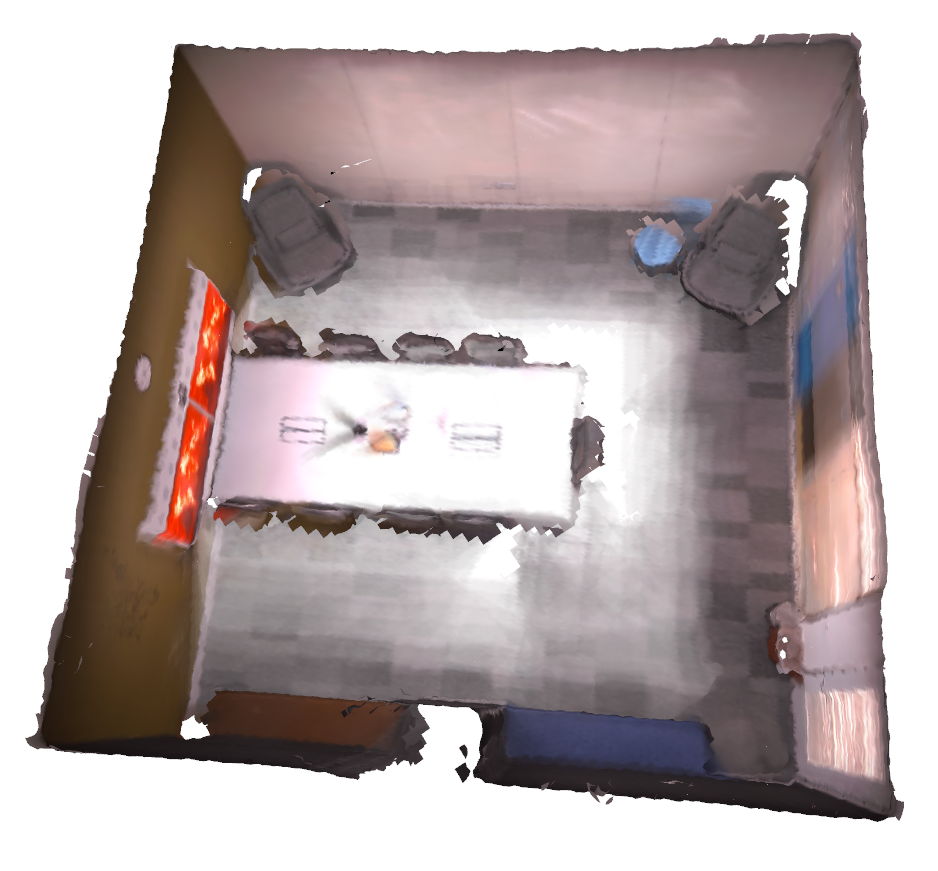}
		\caption{office4}
	\end{subfigure}%
	\begin{subfigure}{.2\textwidth}
		\centering
		\includegraphics[width=1.\linewidth]{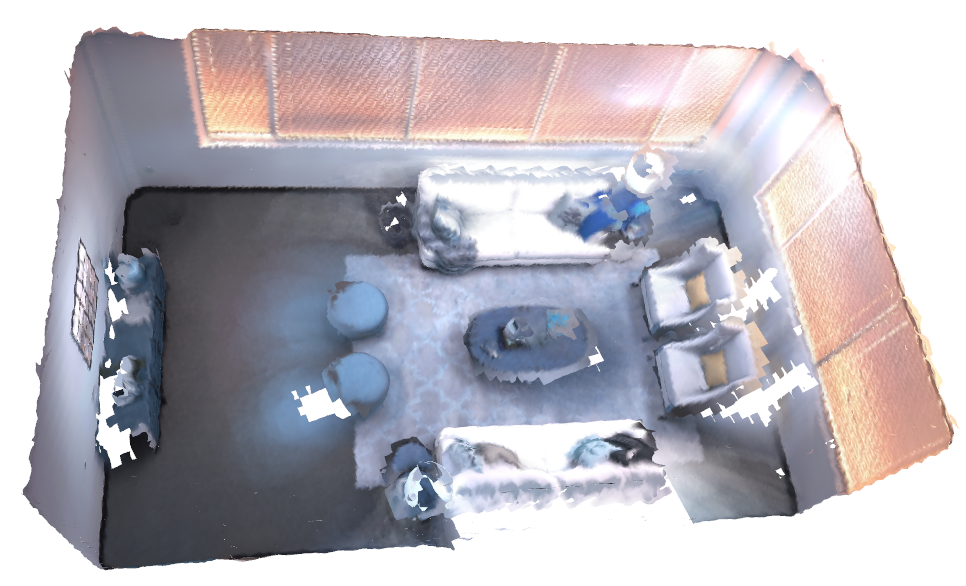}
		\caption{room0}
	\end{subfigure}%
	\begin{subfigure}{.2\textwidth}
		\centering
		\includegraphics[width=1.\linewidth]{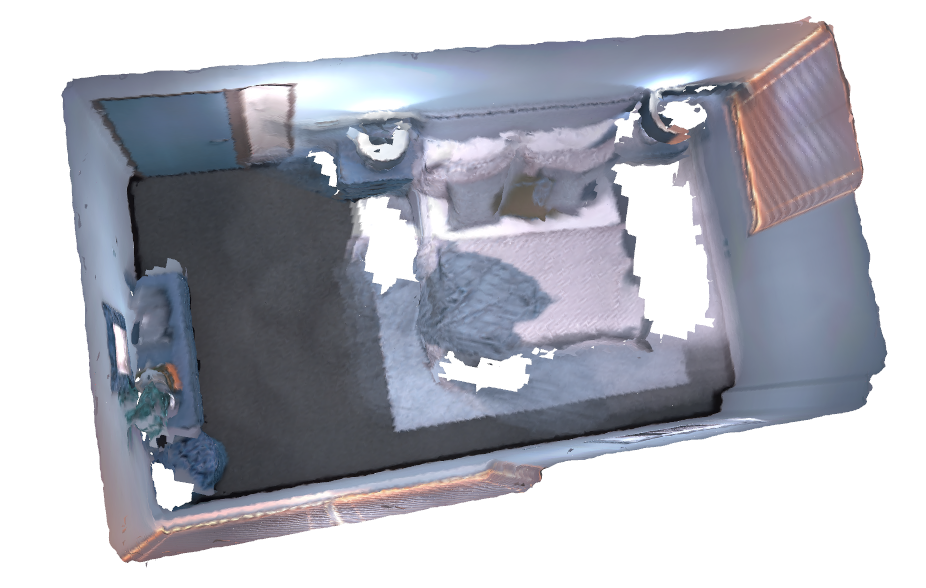}
		\caption{room1}
	\end{subfigure}%
	\begin{subfigure}{.2\textwidth}
		\centering
		\includegraphics[width=1.\linewidth]{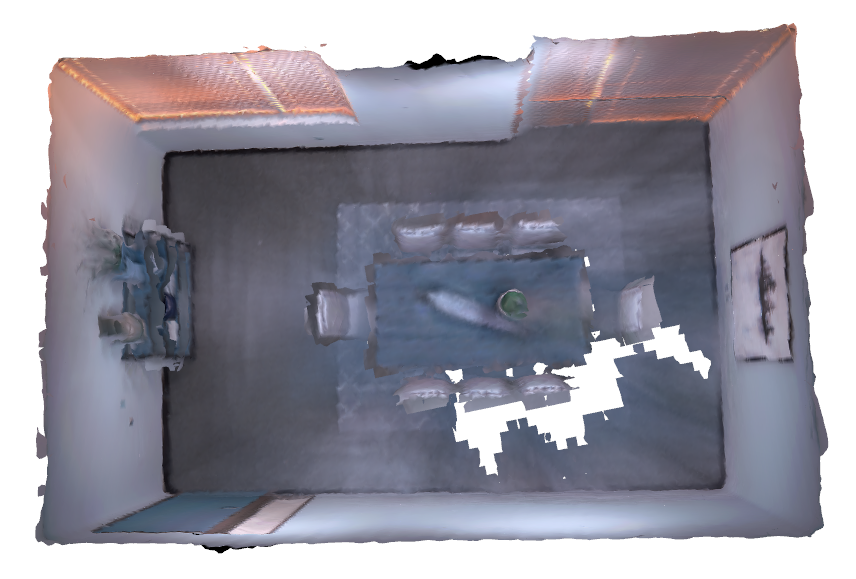}
		\caption{room2}
	\end{subfigure}%
	\caption{Reconstruction Demonstration. Our post-processed textures are averaged from projected image colors.}
	\label{fig:draw_demo}
	\vspace*{-6mm}     
\end{figure*}
%

\subsubsection{Replica Test}

We also evaluate our model on the iMAP~\cite{sucar2021imap} Replica Dataset sequences.
The metrics follow iMAP on accuracy, completion, and completion ratio. 
The completion ratio is an important metric because the ground truth model contains the ceiling which is mostly non-touched in data sequences.
In \cref{tab:replica}, we see that our model scores best on all accuracy tests, and best on most completion and completion ratio.
On the average scores, our model achieves best on accuracy and completion ratio.
Our average completion does not exceed iMAP.

Note that iMAP is a live-training model with differential rendering.
It is naturally capable to complete the blocked region of point clouds.
Our model does not have this point cloud completion function.
This explains why iMAP exceeds ours on completion.
But its higher completion but lower or similar completion ratio of iMAP means the guess of unobserved surface usually fails.

Some results are textured and plotted at \cref{fig:draw_demo}.

\subsection{Efficiency Test}

\subsubsection{Time Cost}
Time efficiency of encoding and remapping directly influences the usage of our model in online reconstruction.
Thus we recorded the time cost of lr-kt3 test from \cref{sec::exp::recons::ICL} that contains a large loop.

For each frame that is fed into the mapping model, it is encoded as a local neural implicit map. The recorded encoding time is $\mathbf{0.0077s\pm0.0019s}$ per frame.

When a loop is detected, certain frames require being removed, transformed and fused again onto the global map. 
This is accomplished with neural implicit maps.

Per frame removal from global map takes $\mathbf{0.0040s\pm0.00038s}$. Per frame transformation and interpolation take $\mathbf{0.018s\pm0.0039s}$. Per frame fusion to global map takes $\mathbf{0.0032\pm0.00044s}$. 

Thus the encoding processes around $\mathbf{130Hz}$, and the remapping processes around $\mathbf{50Hz}$ on our NUC-computer. Therefore this remapping can be well-adequate for the online application.

\subsubsection{Space Cost}
The space cost mainly consists of the Network (encoder, decoder), Neural Implicit Map, and Meshing. 

We evaluate this by saving network parameters and neural implicit maps into files. 
The parameter files are $31.5kB$ for encoder and $207kB$ for decoder. We save full result map from lr-kt3 test with $torch.save$, merely $29.2MB$ are taken. 

During the encoding, we fetched the network parameters that take $26.5kB$. Points are passed to our VNN-Encoder. As we previously provided the specific layers. The space cost is computed as $n\times \max\{1,32,32,9\}\times 3\times2=192n$ $float32$ as an intermediate buffer is not reserved. Thus we count the points into the encoder to compute the encoding buffer $105MB\pm59MB$.

We do not count the space cost of the mesh extracting, as in \cref{fig:pipeline}, it is for visualization and can be operated externally.
\subsection{Demonstration on Campus-scale Reconstruction}
\begin{figure}[!]
	\centering
	\includegraphics[height=.7\linewidth]{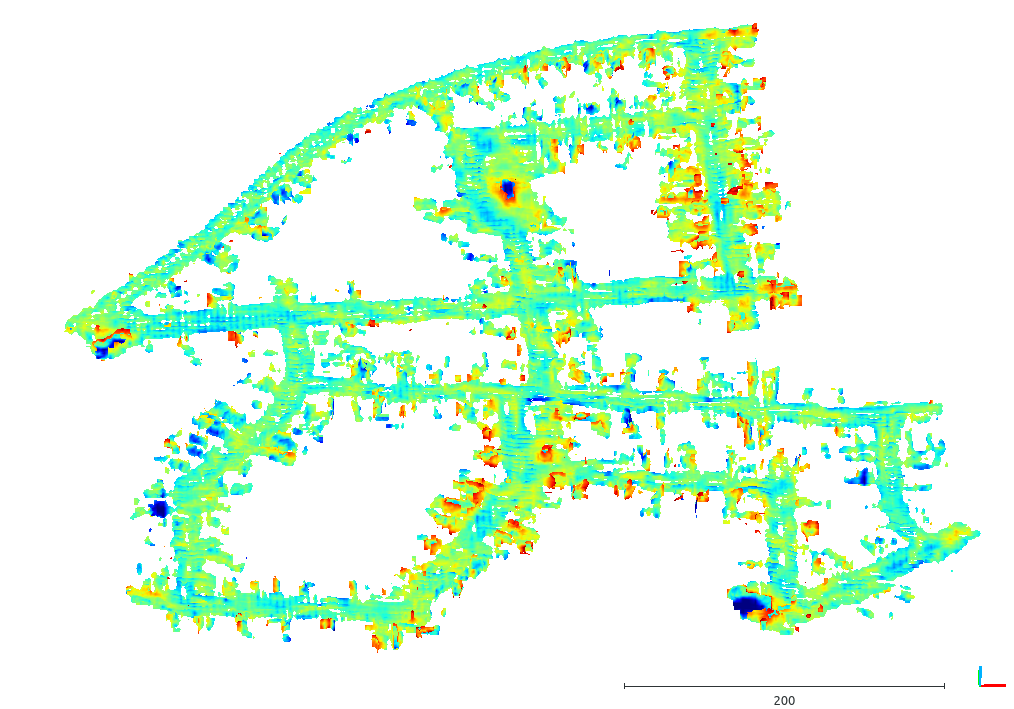}
	\caption{Incremental reconstruction result on KITTI-odometry sequence 00.}
	\label{fig:kitti}
\vspace{-.5cm}
\end{figure}

We are also interested in scenarios that other methods cannot do. We see that for indoor scenes, many sequences do not contain large loops for the front-end restructuring. 
But when it turns to outdoor LiDAR SLAM, such as KITTI-odometry~\cite{geiger2012we}, loop closure shows vital significance to remove the accumulated error for a long trajectory. 
Thus, we attempt to produce a neural implicit mapping on such a scene to further reveal the capability of our algorithm. 
The LiDAR localization model PyICP-SLAM\footnote{\url{https://github.com/gisbi-kim/PyICP-SLAM}} is utilized to provide tracking estimation and the pose graph. Due to the scale difference to the indoor scene and limited available memory, we use a voxel size of $4m$, other hyperparameters are preserved. 
A reconstruction on KITTI-odometry sequence $00$ is given in \cref{fig:kitti}. 

\section{Conclusion}
In this paper, we have presented a neural implicit mapping module that does support loop closing.
By utilizing an SO(3)-equivariant encoder, we are able to implement SE(3)-transformations directly on the neural implicit maps.
In combination with an interpolation step, our mapping module supports updating the neural implicit map when the pose of certain frame changes without touching the original 3D point cloud.
In addition, we showed in our experiments, that our SO(3)-equivariant encoder takes the responsibility of generating neural implicit maps, and based on that, our transforming module functions well and provides high-quality reconstruction with and without a loop closure.
\vspace{-.3cm}
{\small
	\bibliographystyle{IEEEtran}
	\bibliography{ref}
}

\end{document}